\documentclass{article}

\usepackage{microtype}
\usepackage{graphicx}
\usepackage{subfigure}
\usepackage{booktabs} 
\usepackage{xcolor}

\usepackage[toc,page]{appendix}

\usepackage{blindtext}

\usepackage{hyperref}

\usepackage{icml2021}

\icmltitlerunning{BFE and AdaBFE: A New Approach for Stochastic Optimization}

\begin{document}

\twocolumn[
\icmltitle{BFE and AdaBFE: A New Approach in Learning Rate Automation \\ 
for Stochastic Optimization}



\icmlsetsymbol{equal}{*}

\begin{icmlauthorlist}
\icmlauthor{Xin Cao}{equal, to, goo}
\end{icmlauthorlist}

\icmlaffiliation{to}{Georgia Institute of Technology, Atlanta, GA, USA}
\icmlaffiliation{goo}{Georgia , Atlanta, GA, USA}
\icmlaffiliation{ed}{??, ??, ??, ??}

\icmlcorrespondingauthor{Xin Cao}{xincao.tech@gmail.com}

\icmlkeywords{Machine Learning, ICML}

\vskip 0.3in
]



\printAffiliationsAndNotice{\icmlEqualContribution} 

\begin{abstract}
In this paper, a new gradient-based optimization approach by automatically adjusting the learning rate is proposed. This approach can be applied to design non-adaptive learning rate and adaptive learning rate. Firstly, I will introduce the non-adaptive learning rate optimization method: Binary Forward Exploration (BFE), and then the corresponding adaptive per-parameter learning rate method: Adaptive BFE (AdaBFE) is possible to be developed. This approach could be an alternative method to optimize the learning rate based on the stochastic gradient descent (SGD) algorithm besides the current non-adaptive learning rate methods e.g. SGD, momentum, Nesterov and the adaptive learning rate methods e.g. AdaGrad, AdaDelta, Adam... The purpose to develop this approach is not to beat the benchmark of other methods but just to provide a different perspective to optimize the gradient descent method, although some comparative study with previous methods will be made in the following sections. This approach is expected to be heuristic or inspire researchers to improve gradient-based optimization combined with previous methods.
\end{abstract}

\section{Introduction}
\label{submission}

Gradient descent algorithm is one of the most important and widely used optimization method in the area of machine learning and artificial intelligence. During the past few years, with the flourishing progress of machine learning especially of deep learning, many gradient-based optimization methods had been developed in order to improve the performance of traditional gradient descent algorithm (including stochastic gradient descent and mini-batch gradient descent) in the aspects of convergence speed, robustness, the ability of escaping a saddle point. As the basic non-adaptive method, the stochastic gradient descent (SGD) makes all parameters in the model share the same one learning rate in each iteration step. The momentum and Nesterov methods \cite{nesterov1983method} consider the historic gradient information on the basis of SGD, which made the optimization process converges faster and more robustly. However, for high dimensional data, the non-adaptive methods might not efficient or robust enough, this is because in the high dimensional parameter space, the loss function or cost function usually descends with different speed in each dimension, as the gradient in different dimension may largely distinct. This is the motivation why the adaptive learning rate methods were developed based on these non-adaptive methods I mentioned above, and many of the adaptive methods inherited concepts such as momentum, Nesterov gradients from the non-adaptive ones. For instance, recent adaptive gradient descent algorithms include AdaGrad \cite{duchi2011adaptive}, Adadelta \cite{zeiler2012adadelta}, RMSProp \cite{tieleman2012lecture}, Adam \cite{kingma2014adam}, and many other first-order gradient descent algorithms are modified based on these methods. In this paper, I will introduce a new method that automatically adjusts the learning rate by exploring the forward information of the loss function instead of using the historic information, which is called Binary Forward Exploration (BFE) algorithm. I will describe how the BFE method works and then discuss potential adaptive versions of BFE (AdaBFE). With the development of AI-Science area, the strategy of this algorithm has potential to be applied to traditional scientific problems as well, which are usually based on modeling or analysis of the scientific data from space exploration mission \cite{cao2022machine, chu2021electrostatic, chu2021dayside, liu2012dipole, liuzzo2021investigating}.

\section{Binary Forward Exploration Algorithm}

The Binary Forward Exploration method was designed to compare the future loss functions by calculating an original learning rate and a binary reduced or multiplied learning rate.

\subsection{Algorithms}

The Binary Forward Exploration method was originally modified from and improved based on one of the finite-difference methods, which has been applied on the solving differential equations for the computational physics problems. For instance, a test particle numerical simulation has utilized a similar method to model the dynamics of charged particles in the space environment, with a quasi-adiabatic process \cite{Cao2013Trajectorymethodof3Dtestparticles}.

\begin{algorithm}[H]
   \caption{Binary Forward Exploration (BFE), the proposed algorithm in non-adaptive learning rate automation for stochastic optimization. See the text for details. Loss function is denoted as $Loss=f(\theta)$. Default setting for error limit ratio for Binary Detection Learning Rate is $\epsilon = 0.001$, meaning one thousandth.}
   \label{alg:example}
\begin{algorithmic}
   \STATE Initialize learning rate $\eta$ (e.g. 0.001)
   \STATE Initialize $\epsilon_v$ (e.g. 0.001)
   \STATE Initialize $\epsilon_c > \epsilon_v$
   \STATE Initialize parameter vector $\theta_0$
   \STATE Initialize time-step $t=0$
   \WHILE{$\theta_t$ not converged}
     \STATE $t=t+1$
     \IF{$\epsilon_c \geq \epsilon_v$}
       \WHILE{$\epsilon_c \geq \epsilon_v$}
         \STATE $\theta^{*}_{t} = \theta_t - \eta \frac{\partial f(\theta)}{\partial \theta_{t}}$
         \STATE $\theta^{+}_{t} = \theta_t - \frac{\eta}{2} \frac{\partial f(\theta)}{\partial \theta_{t}}$
         \STATE $\theta^{'}_{t} = \theta^{+}_{t} - \frac{\eta}{2} \frac{\partial f(\theta)}{\partial \theta^{+}_{t}}$
         \STATE $Loss1 = [f(\theta)]_{\theta^{*}_{t}}$ $ $ $ $ (loss value at $\theta^{*}_{t}$)
         \STATE $Loss2 = [f(\theta)]_{\theta^{'}_{t}}$ $ $ $ $ (loss value at $\theta^{'}_{t}$)
         \STATE $\epsilon_c = |Loss2-Loss1|$
         \STATE $\epsilon_v= 0.5 \cdot (|Loss2|+|Loss1|) \cdot \epsilon$ or $min(|Loss2|\cdot \epsilon, |Loss1|\cdot \epsilon)$ or any other predefined factor or functions, e.g. decay with epochs
         \STATE $\eta = \frac{\eta}{2}$
       \ENDWHILE
       \STATE $\theta_{t} = \theta^{+}_{t}$
     \ELSE
       \WHILE{$\epsilon_c < \epsilon_v$}
         \STATE $\theta^{+}_{t} = \theta_t - \eta \frac{\partial f(\theta)}{\partial \theta_{t}}$
         \STATE $\theta^{'}_{t} = \theta^{+}_t - \eta \frac{\partial f(\theta)}{\partial \theta^{+}_{t}}$
         \STATE $\theta^{*}_{t} = \theta_{t} - 2\eta \frac{\partial f(\theta)}{\partial \theta_{t}}$
         \STATE $Loss1 = [f(\theta)]_{\theta^{'}_{t}}$ $ $ $ $ (loss value at $\theta^{'}_{t}$)
         \STATE $Loss2 = [f(\theta)]_{\theta^{*}_{t}}$ $ $ $ $ (loss value at $\theta^{*}_{t}$)
         
         \STATE $\epsilon_c = |Loss2-Loss1|$
         \STATE $\epsilon_v= 0.5 \cdot (|Loss2|+|Loss1|) \cdot \epsilon$ or $min(|Loss2|\cdot \epsilon, |Loss1|\cdot \epsilon)$ or any other factor or functions
         \STATE $\eta = 2\eta$
       \ENDWHILE
       \STATE $\eta = \frac{\eta}{2}$
       \STATE $\theta_{t} = \theta^{+}_{t}$
     \ENDIF
   \ENDWHILE
   \RETURN $\theta_{t}$ (Resulting Optimized Parameters)
   
\end{algorithmic}
\end{algorithm}

See algorithm 1 for pseudo-code of the proposed algorithm, $\frac{\partial f(\theta)}{\partial \theta_{t}}$ is the gradient of loss function at $\theta_{t}$, and $\frac{\partial f(\theta)}{\partial \theta^{+}_{t}}$ is the gradient of loss function at $\theta^{+}$. In this article, the gradient expression $\frac{\partial f(\theta)}{\partial \theta_{t}}$ is equivalent to $\nabla f(\theta_{t})$, where $\theta_{t}$ represents the parameter vectors or tensors at time-step $t$, and the loss function w.r.t the parameter is $f(\theta)$.

This algorithm compares the values of $\epsilon_{comp}$ and $\epsilon_{val}$ at each timestep ($\epsilon_{c}$ and $\epsilon_{v}$ for short in the algorithm's pseudo-code), where $\epsilon_{comp}$ can be considered as the inaccuracy or relative error of one-time updated loss function, and $\epsilon_{val}$ is the threshold value that decides if the iteration of finding a “good enough” learning rate should stop at each time-step. These variables will be defined in the following description.

\subsection{BFE’s update rule:}

\begin{figure}[ht]
\vskip 0.2in
\begin{center}
\centerline{\includegraphics[width=\columnwidth]{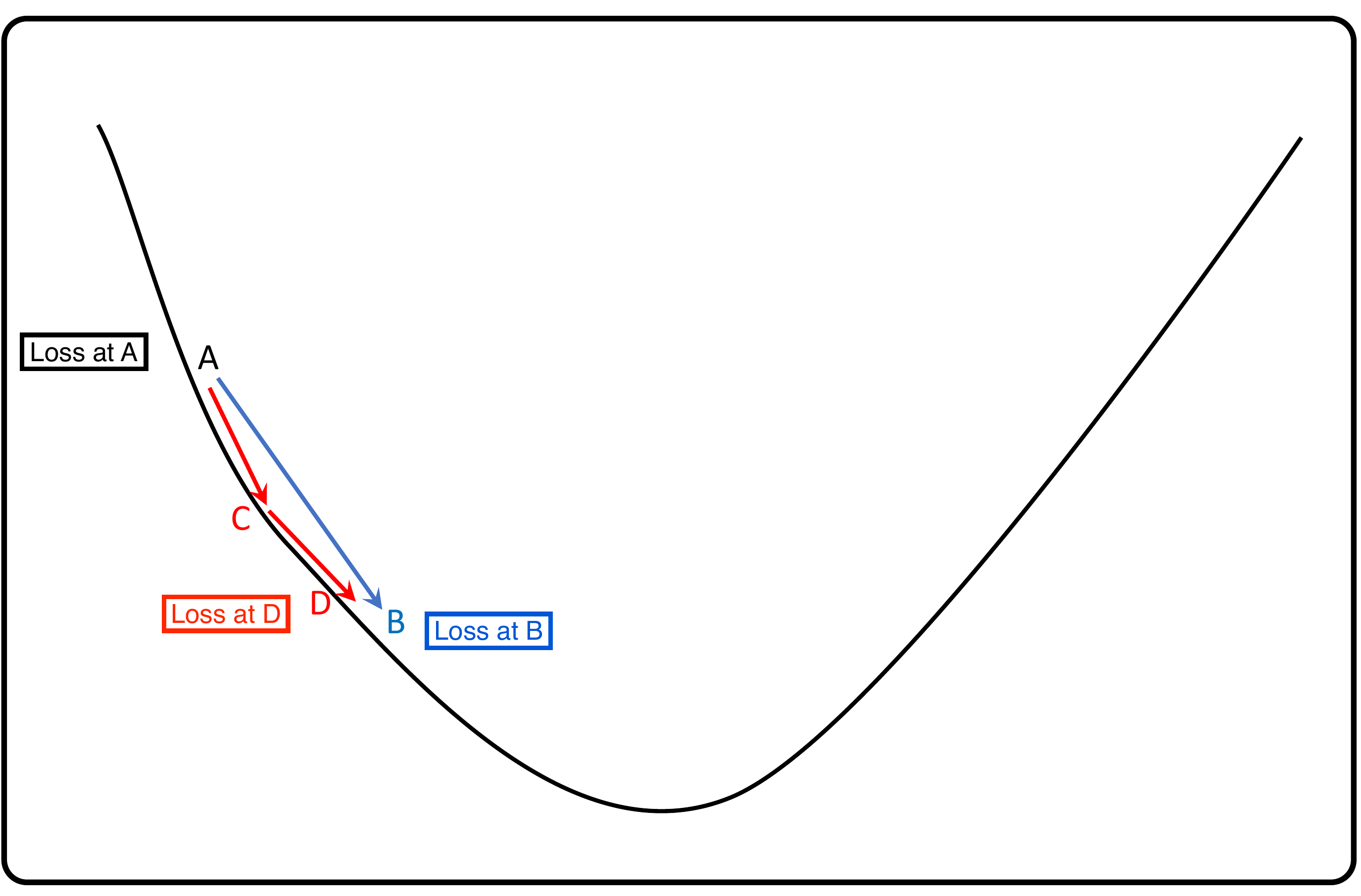}}
\caption{The update rule of BFE algorithm}
\label{icml-historical}
\end{center}
\vskip -0.2in
\end{figure}

As the algorithm 1 shows, $\epsilon_{comp}$ can be calculated by the absolute value of the subtraction between $Loss1$ (LS1) and $Loss2$ (LS2) at the current time-step. In the “while $\epsilon_{comp} \geq \epsilon_{val}$ do” loop,  $Loss1$ is defined to be the loss function value at $\theta^{*}_{t}$, where $\theta^{*}_{t}$ is updated from the parameter $\theta_{t}$ by using the gradient descent method with the current learning rate and gradient at $\theta_{t}$. In contrast, $Loss2$ is defined as follow: at first, update the parameter by using a half of the current learning rate and the gradient at $\theta_{t}$ to obtain $\theta^{+}_{t}$, and then repeat to update it by using the current learning rate again and the gradient at $\theta^{+}_{t}$ to obtain $\theta^{'}_{t}$ (Please note: $\theta^{'}_{t}$ in this article is not the derivative of $\theta_{t}$). Therefore, we can calculate the value of loss function $\theta^{'}_{t}$, which is $Loss2$. The absolute value of the difference of $Loss1$ and $Loss2$ represents how large the two procedures which correspond to $Loss1$ and $Loss2$ differ from each other. $\epsilon_{val}$ is used to identify if $\epsilon_{comp}$ is small enough such that the corresponding learning rate can be considered to be safe for the parameter updating during current time-step. $\epsilon_{val}$ can be defined as needed, for instance, it can be defined as $\frac{\mid Loss1 \mid + \mid Loss2 \mid}{2}\cdot\epsilon$ or $ 0.5 \cdot (\mid Loss1 \mid + \mid Loss2 \mid) \cdot\epsilon$, where $\epsilon$ is the error limit ratio, which is a fraction, e.g. 0.001. $\epsilon_{val}$ can be also defined as a function of training epochs, e.g. a decreasing $\epsilon_{val}$ with the increasing epochs, which makes $\epsilon_{val}$ as a decay factor function. The $\epsilon_{comp}$ means a fraction of the mean of the sum of two loss values or $\min \left(\mid Loss1\mid \cdot \epsilon,\space \mid Loss2\mid \cdot \epsilon\right)$, or other definition such as even a constant number. During the current “while $\epsilon_{comp} \geq \epsilon_{val}$ do” loop, update the learning rate to be half of the original rate, and repeat the same procedure if $\epsilon_{comp} \geq \epsilon_{val}$ is still satisfied, until $\epsilon_{comp} < \epsilon_{val}$ occurs when the learning rate is set to the final learning rate for the current time-step $t$.

In the “while $\epsilon_{comp}<\epsilon_{val}$ do” loop, $Loss1$ is defined as follow: at first, update the parameter by using the current learning rate and the gradient at $\theta_{t}$ to obtain $\theta^{+}_{t}$, and then repeat to update it by using a half of the current learning rate and the gradient at $\theta^{+}_{t}$ to obtain $\theta^{'}_{t}$. In contrast, $Loss2$ is defined to be the loss function value at $\theta^{\star}_{t}$, where $\theta^{\star}_{t}$ is updated from the parameter $\theta_{t}$ by using the gradient descent method with twice of the current learning rate and gradient at $\theta_{t}$. Similarly, during the current “while $\epsilon_{comp}<\epsilon_{val}$ do” loop, update the learning rate to be twice of the original rate, and repeat the same procedure if $\epsilon_{comp}<\epsilon_{val}$ is still satisfied, until $\epsilon_{comp} \geq \epsilon_{val}$ occurs when the learning rate is set to the final learning rate for the current time-step $t$.

Figure 1 shows the sketch of how the learning rate updates in a single step. If the loss decrease after processing one step (as blue arrow $A\rightarrow B$ shows) via the original learning rate is comparable to that after processing two continuous steps (as the red arrows $A\rightarrow C\rightarrow D$ show) via half of the original learning rate, it represents that the original learning rate is suitable for the current time-step. If the difference between the two losses exceeds the pre-defined threshold value, the learning rate will be decreased to half of it and compare the re-calculated losses in the next iteration, until the difference of the two losses in that iteration is less than or equal to the threshold value. If the difference of two calculated losses is less than the threshold value in the beginning of the current iteration, the learning rate will be doubled (If the update $A\rightarrow C$ is the original, and $A\rightarrow B$ is the doubled for this situation; The updated loss after processing one step via twice of the original learning rate will be used to compared to that after processing two continuous steps via the original learning rate) during the iterations until a significant difference of the losses is approached, and then the learning rate will be cut by half of it just once to guarantee the learning rate is not excessive. The recurrence of such updating the learning rate occurs during every iteration until the local optima is approached and the parameters are converged.

After calculating the learning rate and updating the parameters through the “while $\epsilon_{comp} \geq \epsilon_{val}$ do”  and the “while $\epsilon_{comp}<\epsilon_{val}$ do” loop for each time iteration, the time-step is then updated $t = t + 1$, until the parameters are converged to a local optimum, e.g. the gradient approaches a small threshold number. For instance, the non-convergence condition can be $\frac{\partial f(\theta)}{\partial \theta_{t}}>lim{\_}zero$, where $lim{\_}zero$ is the zero approaching critical value (e.g. 0.001) for the loss function. In general, the inner loops determine the learning rate by a binary factor for each time step, and the outer loop determines when the model parameters approaches a local optimum and then stop to update. 

\section{Experiments}

In order to evaluate the performance of the method, the linear regression model was used to investigate the regression task. The result shows that BFE can add a constraint of the artificial upper limit of current step’s learning rate.

The proposed method in this paper is evaluated by using a typical linear regression model, the corresponding loss function of which is therefore a quadratic function form. The quadratic function is efficient to observe the convergence behavior and also meet the characteristics of a local minimum \cite{lucas2018aggregated, sutskever2013importance, o2015adaptive}. However, a more complex distribution of the loss in a multi-dimensional parameter space by using the neural network model and different datasets such as CIFAR-10, CIFAR-100 \cite{krizhevsky2009learning}, MNIST \cite{lecun1998gradient} and Mini-ImageNet \cite{vinyals2016matching} will be investigated in the future work, which is ideal to compare a variety of optimizers with guarantee of approaching local minimum.

Without any generality, the data can be sampled from a linear equation such as:
$Y=W_0 X+b_0+\epsilon_0$,
where $Y$ is target and $X$ is feature, and $\epsilon_0$ is the noise term, and $W_0$ and $b_0$ is a constant value e.g. $W_0=5$ and $b_0=9$. Correspondingly, a linear regression model can be written as:
$Y=WX+b$,
where $W$ and $b$ are the parameters of weight and bias term which are assumed known and will be optimized by using the proposed algorithm later.

\begin{figure}[ht]
\vskip 0.2in
\begin{center}
\centerline{\includegraphics[width=\columnwidth]{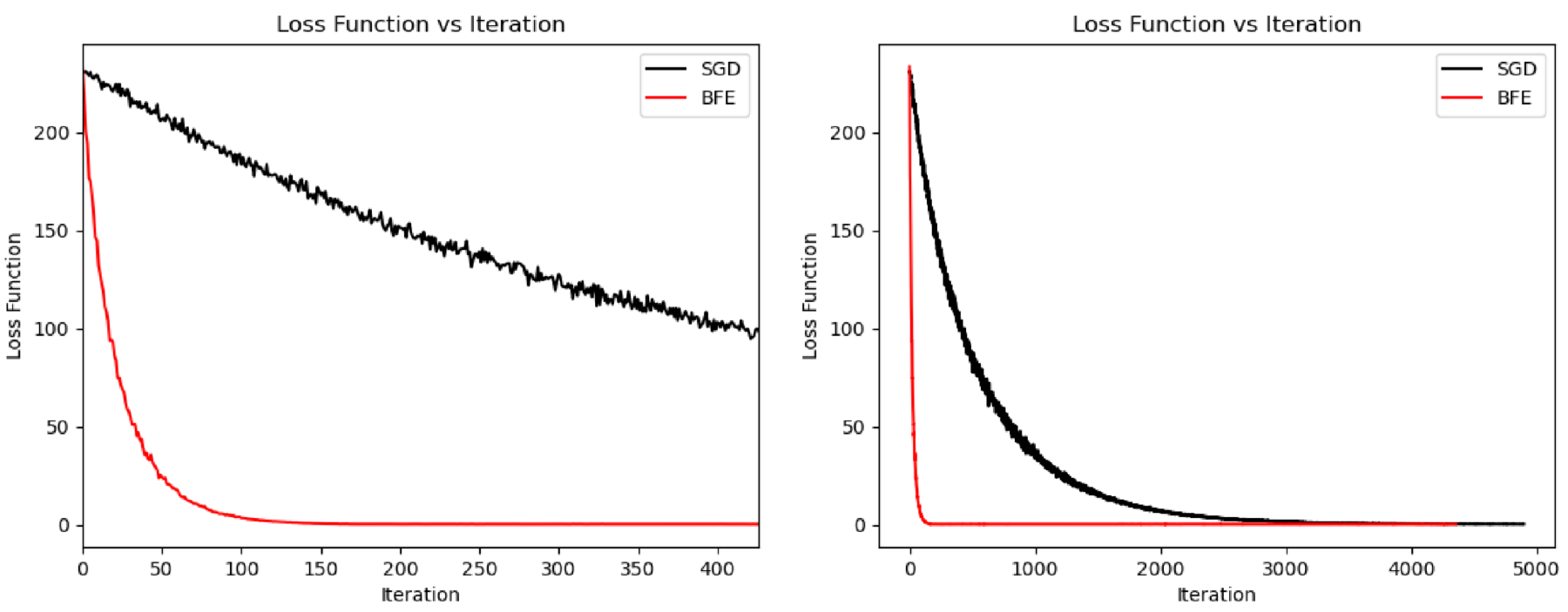}}
\caption{The variation of the loss with the iteration over the whole dataset. The red curve represents the loss decrease by using the BFE optimizer and the black represents that by using the mini-batch gradient descent algorithm (without momentum or Nesterov terms).}
\label{icml-historical}
\end{center}
\vskip -0.2in
\end{figure}

As Figure 2 shows, the BFE algorithm made the loss function converge much faster than the mini-batch gradient descent method, both of which are non-adaptive algorithm. The batch size is set to be 512 in the experiments. Figure 2 demonstrated the loss function values decrease with the increasing iterations. The left panel shows the situation from start to less than $500^{th}$ iteration, and the right panel shows the situation from start to around $5000^{th}$ iteration. The local minimum is approached at around $100^{th}$ iteration by using BFE and at over $2500^{th}$ iteration by using the SGD algorithm. Especially during the very beginning, the BFE algorithm can make the loss function values decrease much faster than the SGD algorithm.

\begin{figure}[ht]
\vskip 0.2in
\begin{center}
\centerline{\includegraphics[width=\columnwidth]{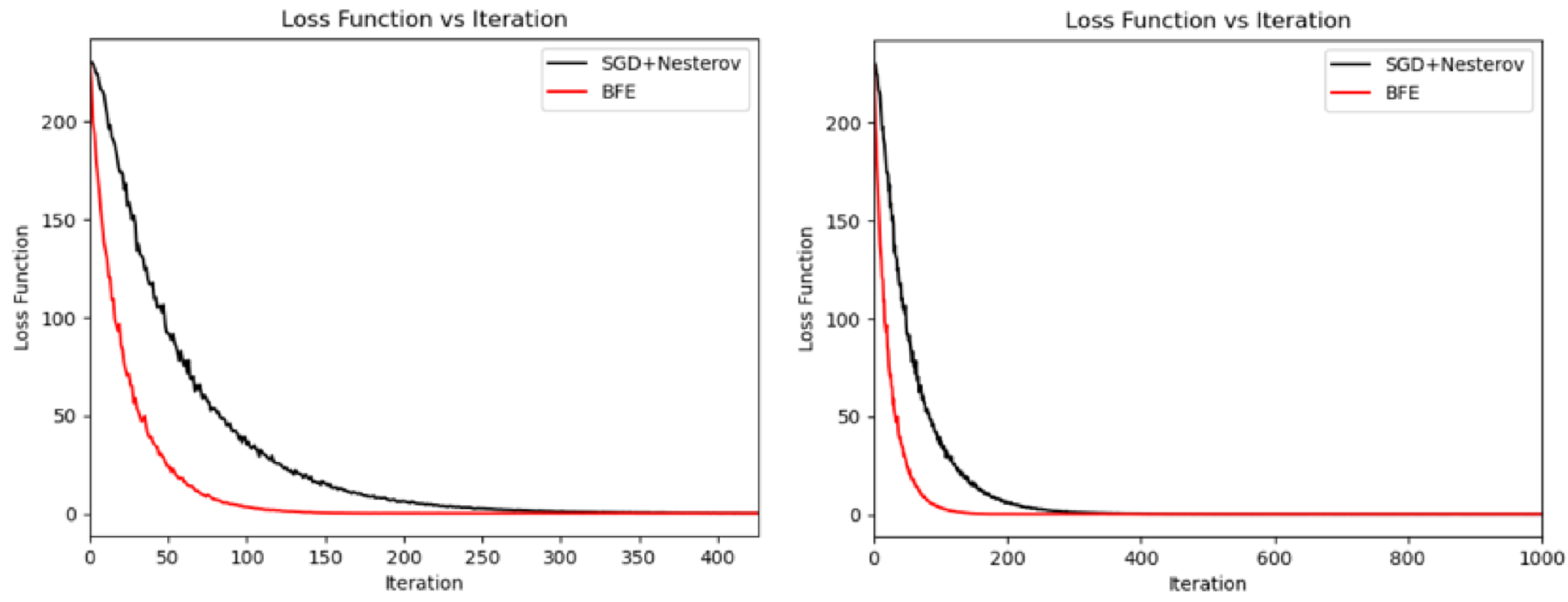}}
\caption{The variation of the loss with the iteration, which shows the comparison between the SGD+ Nesterov momentum and BFE.}
\label{icml-historical}
\end{center}
\vskip -0.2in
\end{figure}

The SGD with Nesterov momentum algorithm utilized the previous historic gradient information when computing the gradient at current time step, which can help the loss function converges faster than the SGD without Nesterov momentum. The SGD with Nesterov momentum algorithm is shown as below:

\begin{equation}
\begin{array}{c}
v_i=\beta v_{i-1}+g(w_{i-1}-\alpha\beta v_{i-1})\\ w_i=w_{i-1}-\alpha v_i
\end{array}
\end{equation}

where beta in the equation represents the weight or ratio of the previous history information applied to the current step, and the default value of beta is normally set to be 0.9. By changing beta value, how the historic information affects the current parameter updates can be modulated. Although the Nesterov momentum boosted the convergence faster than the original SGD, it is still slower than the BFE.

\begin{figure}[ht]
\vskip 0.2in
\begin{center}
\centerline{\includegraphics[width=\columnwidth]{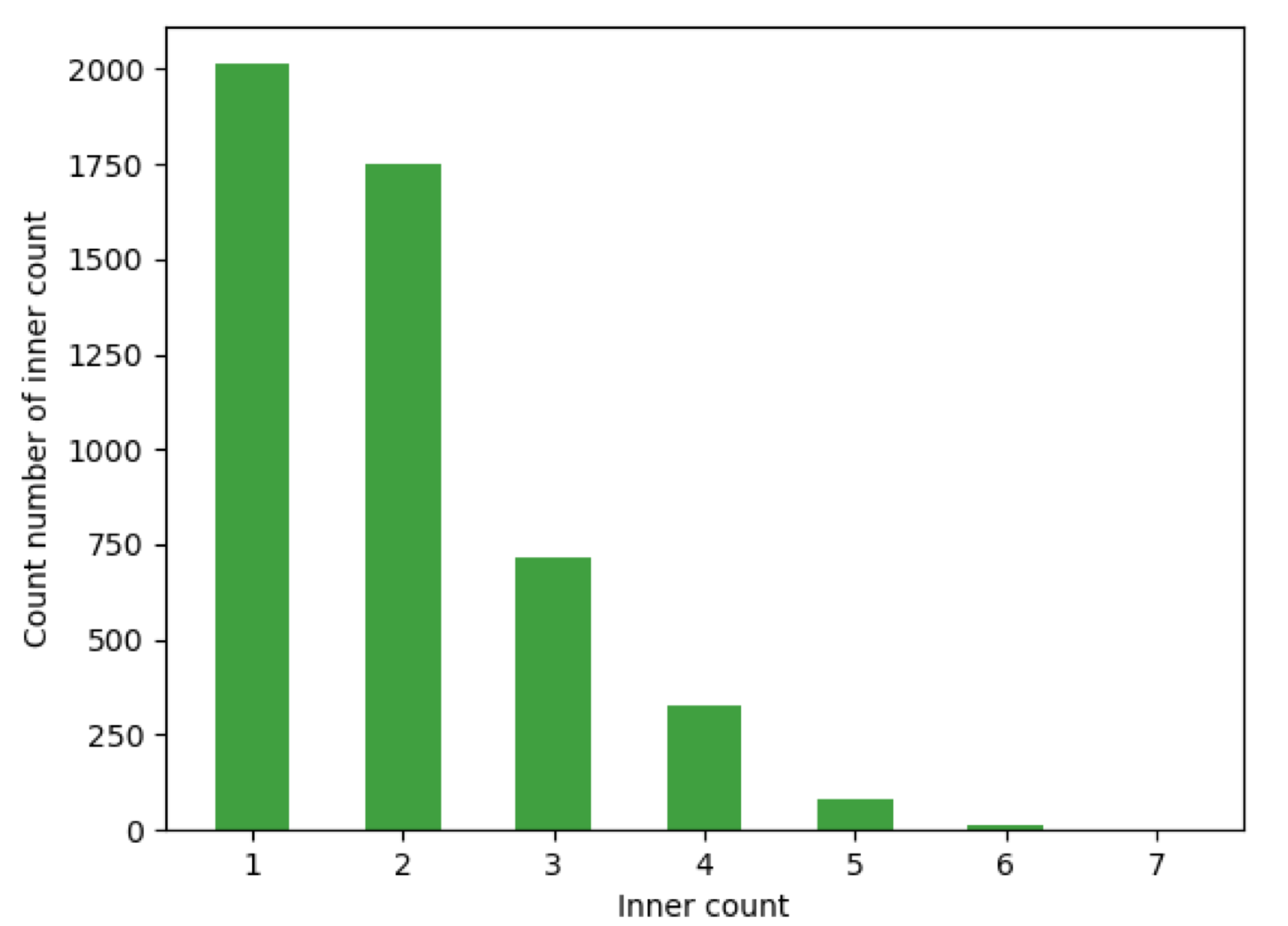}}
\caption{The statistics about the number of inner loops and its corresponding count number. The average number of inner loops of BFE optimizer is 1.93.}
\label{icml-historical}
\end{center}
\vskip -0.2in
\end{figure}

In order to investigate the time complexity of the BFE algorithm, the number of inner loops (zoom-in / zoom-out process) has been recorded during the optimization. The majority concentrated to be 1 or 2 loops, and the average inner loop number is about 1.93, which means on average the inner loop would be executed about twice per time step’s update.

\begin{figure}[ht]
\vskip 0.2in
\begin{center}
\centerline{\includegraphics[width=\columnwidth]{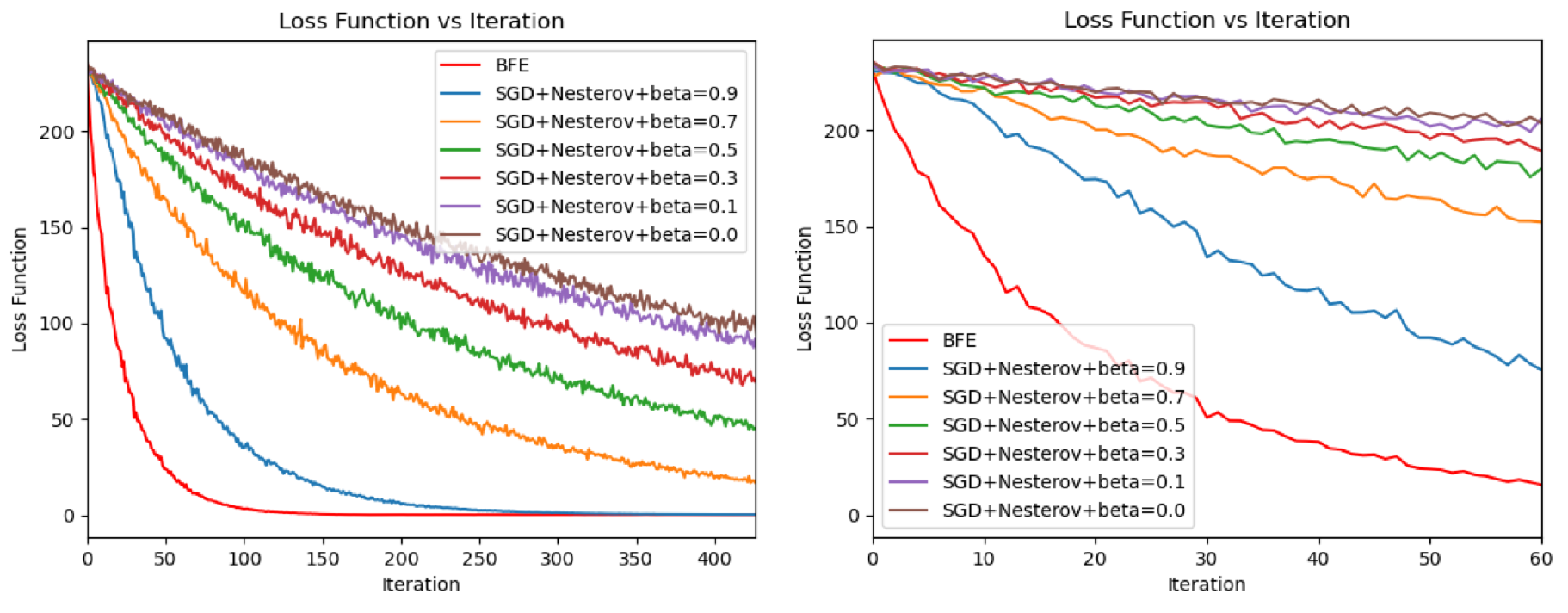}}
\caption{The loss vs iterations over the whole dataset.}
\label{icml-historical}
\end{center}
\vskip -0.2in
\end{figure}

Figure 5 reveals the comparison between the BFE algorithm and the SGD with Nesterov momentum with different beta value. The beta value is respectively set to be 0.9, 0.7, 0.5, 0.3, 0.1 and 0.0, which shows when the beta is smaller, the convergence speed is slower. During the very beginning of the optimization (as shown as the right panel), the BFE algorithm is capable of steeply decrease to slide along with the ambient gradient from the mini-batch data. In contrast, the SGD with Nesterov momentum with exponentially decaying moving average of previous gradients can not well capture the initial variation of the gradient or the second order variation of the loss function, which is because the shortage of historic information in the very beginning makes the algorithm insufficient to obtain a good enough learning rate. In practice, an extra warm-up strategy is needed to artificially adjust the learning rate, since without it, the optimization probably leads to a non-optimal local minimum. However, the BFE algorithm have the privilege to skip using any other warm-up strategy since it generally has better performance on tracking the gradient variation of the forward steps.

\begin{figure}[ht]
\vskip 0.2in
\begin{center}
\centerline{\includegraphics[width=\columnwidth]{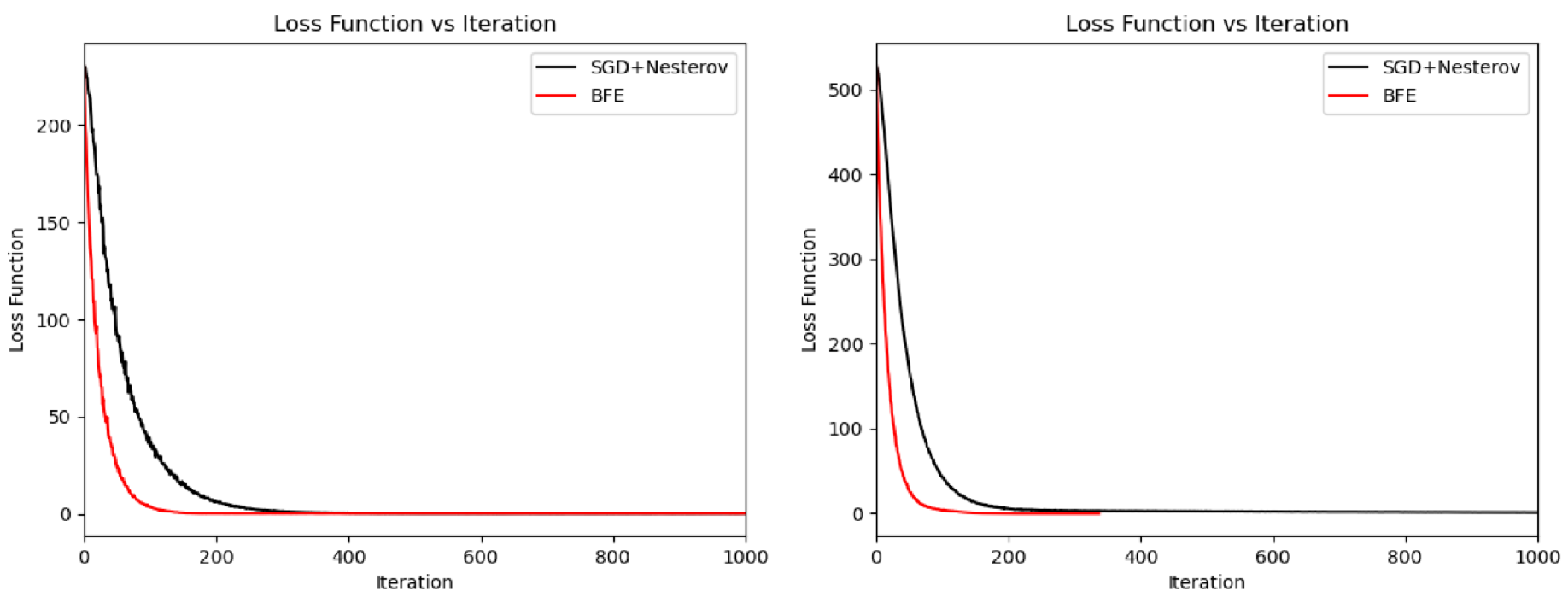}}
\caption{The loss variation with iterations over the whole dataset after data was normalized. The left and right panels reveal the results by using the unnormalized and normalized data. The BFE application to the normalized data has faster convergence.}
\label{icml-historical}
\end{center}
\vskip -0.2in
\end{figure}

Figure 6 compared the convergence speed of BFE and SGD with Nesterov momentum  by respectively using unnomalized and normalized data, which shows that compared with SGD with Nesterov momentum, the BFE algorithm can boost the convergence to a local minimum if applied to the normalized data. The benefit of BFE method is that it can automatically find out an optimum learning rate for the batch sized data during each step.

Besides tracking the variation of loss decrease by adjusting the learning rate, we can also track the gradient variation of the loss function rather than the variation of loss decrease itself. Both of the two metrics essentially explore forward the loss landscape and enable to probe the second-order information of it via the relatively low computational complexity which approximates that of the first-order algorithms. In the following section, the adaptive version of BFE algorithm will be introduced. This adaptive algorithm is designed to adjust different learning rate in each parameter dimension based on the BFE method we discussed above.

\section{AdaBFE Algorithm}

The Adaptive Binary Forward Exploration method was an adaptive extension to BFE algorithm by adjusting different learning rate per parameter’s dimension via tracking the corresponding loss decrease, or via tracking gradient variation of the loss function along each dimension in the parameter space.
As mentioned in the previous section, both of non-adaptive and adaptive algorithm track the gradient of the loss function in each dimension, but the adaptive method calculates respective learning rate in each dimension in the parameter space, compared with the non-adaptive method with a global learning rates among all dimensions.

\subsection{BFE of gradient change}

\begin{algorithm}[H]
   \caption{BFE of gradient change, the proposed algorithm in non-adaptive learning rate automation for stochastic optimization. See the text for details. We express Loss Function as $Loss=f(\theta)$. Default setting for error limit ratio for Binary Detection Learning Rate is $\epsilon = 0.001$, meaning one thousandth.}
   \label{alg:example}
\begin{algorithmic}
   \STATE Initialize learning rate $\eta$ (e.g. 0.001)
   \STATE Initialize $\epsilon_v$ (e.g. 0.001)
   \STATE Initialize $\epsilon_c > \epsilon_v$
   \STATE Initialize parameter vector $\theta_0$
   \STATE Initialize time-step $t=0$
   \WHILE{$\theta_t$ not converged}
     \STATE $t=t+1$
     \IF{$\epsilon_c \geq \epsilon_v$}
       \WHILE{$\epsilon_c \geq \epsilon_v$}
         \STATE $g_i = \frac{\partial f(\theta)}{\partial \theta_{t,i}}$
         \STATE $\theta^{*}_{t,i} = \theta_{t,i} - \eta \cdot g_i$
         \STATE $g^{*}_i = \frac{\partial f(\theta)}{\partial \theta^{*}_{t,i}} $
         \STATE $\epsilon_i = arctan(abs((g^*_i-g_i)/(1+g^*_i \cdot g_i)))$
         \STATE $\epsilon_c = max(\epsilon_i)$
         \STATE $\epsilon_v= 1$ or any other values
         \STATE $\eta = \frac{\eta}{2}$
       \ENDWHILE
       \STATE $\eta = 2\eta$
       \STATE $\theta_{t,i} = \theta^{*}_{t,i}$
     \ELSE
       \WHILE{$\epsilon_c < \epsilon_v$}
         
         \STATE $g_i = \frac{\partial f(\theta)}{\partial \theta_{t,i}}$
         \STATE $\theta^{*}_{t,i} = \theta_{t,i} - \eta \cdot g_i$
         \STATE $g^{*}_i = \frac{\partial f(\theta)}{\partial \theta^{*}_{t,i}} $
         \STATE $\epsilon_i = arctan(abs((g^*_i-g_i)/(1+g^*_i \cdot g_i)))$
         \STATE $\epsilon_c = max(\epsilon_i)$
         \STATE $\epsilon_v= 1$ or any other values
         \STATE $\eta = 2\eta$
       \ENDWHILE
       \STATE $\eta = \frac{\eta}{2}$
       \STATE $\theta_{t,i} = \theta^{*}_{t,i}$
     \ENDIF
   \ENDWHILE
   \RETURN $\theta_{t}$ (Resulting Optimized Parameters)
\end{algorithmic}
\end{algorithm}

The gradient change version of BFE method simply tracks the gradient change of the loss function, rather than tracking the decrease of loss itself, which enables to extend the algorithm to non-adaptive easily. As algorithm 2 shows, during each step, the new gradient will be calculated after the weights are updated through the calculated learning rate from previous step. We then compare the gradient change between the newly calculated gradient $g^*_i$ and the current gradient $g_i$ via computing the angular variation of the gradient during this new update. After that, we then get the maximum angular variation among all dimensions so that we can compare it with the pre-defined threshold angle $epsilon_{v}$ (e.g. 1 degree), in order to decide if the current loop needs to be continued or jumped out. In the “while $\epsilon_{comp} \geq \epsilon_{val}$ do” loop, the learning rate will be updated to be half of it, and in the “while $\epsilon_{comp} < \epsilon_{val}$ do” loop, the learning rate will be doubled. During this iteration, the learning rate is updated via such a zoom-in-zoom-out process. After each time-step, the parameter $\theta_{t,i}$ is thus updated. The whole optimization process will be terminated until a local loss minimum is approached.

\begin{figure}[ht]
\vskip 0.2in
\begin{center}
\centerline{\includegraphics[width=\columnwidth]{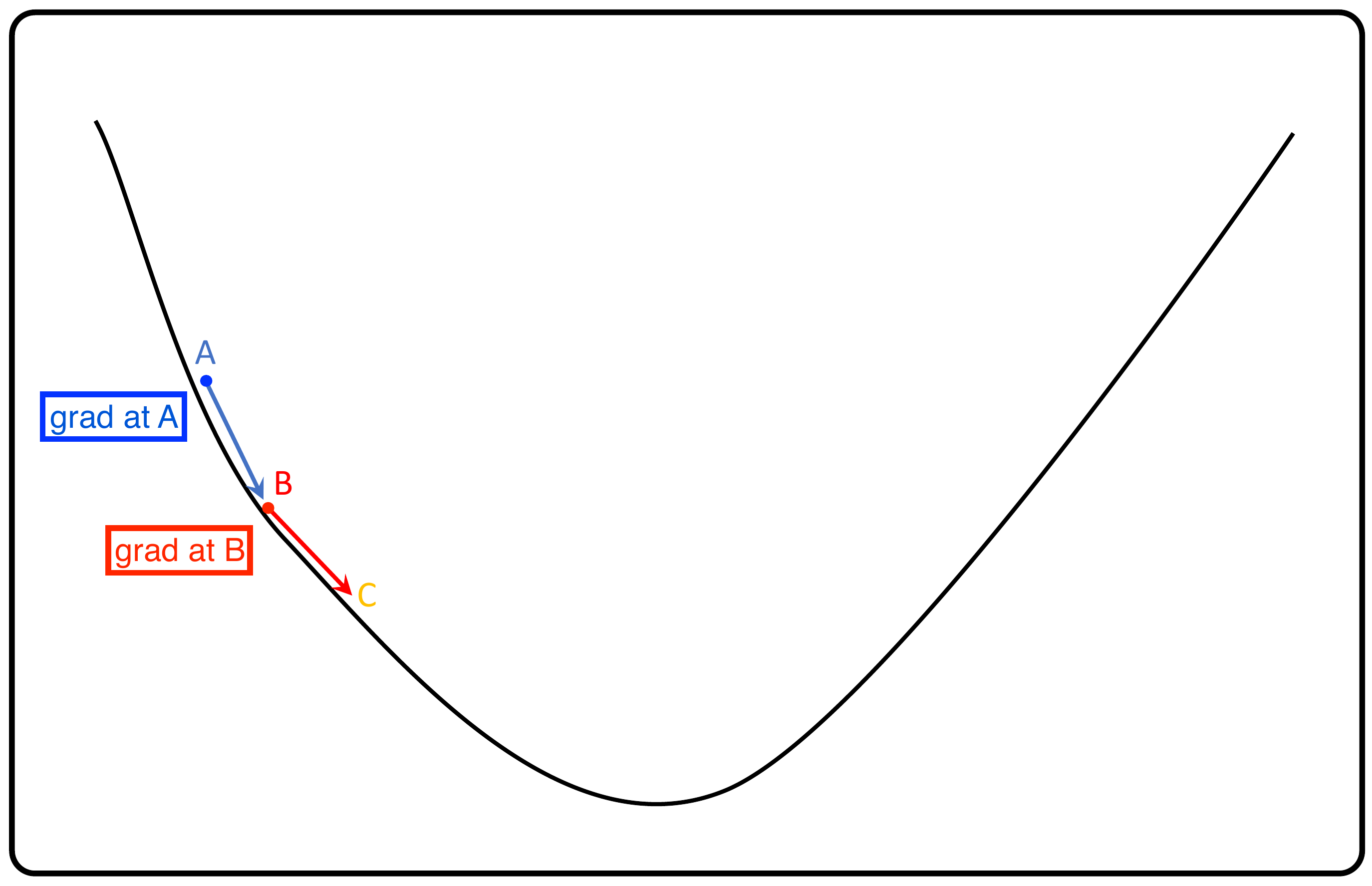}}
\caption{The update rule of BFE of gradient change algorithm}
\label{icml-historical}
\end{center}
\vskip -0.2in
\end{figure}

To better demonstrate how the BFE of gradient change works, Figure 7 shows the update process of BFE of gradient change, which displays how the learning rate updates during a single step. We consider that (1) the updated gradient (gradient a point B) after processing one step (as red arrow $B\rightarrow C$ shows) via a learning rate, and (2) the previous gradient (gradient a point A) before the parameter is updated (as the blue arrows $A\rightarrow B$ show). If the change of two sequential gradients is larger than the pre-defined threshold value (e.g. an absolute value such as 1 degree, or a relative value such as $0.01 \times$ the tilted angle of the gradient at point A), the learning rate should be decreased to half of it, and repeat the re-calculation of the new sequential gradients in the next iteration, until the difference of the two gradients in that iteration is less than or equal to the threshold value. If the difference of two calculated gradients is less than the threshold value in the beginning of the first iteration, the learning rate will be doubled and repeats the same process of comparing the two gradients during each iterations until a significant difference of the gradients is approached, and then the learning rate will be cut by a half just once to guarantee the learning rate is not excessive. The recurrence of such updating the learning rate occurs during every iteration until the local optima is approached and the parameters are thus converged. The iterative process is demonstrated in Figure 8. Correspondingly, the iterative process of original BFE (not shown here) is similar to Figure 8, but just replace every step in it with the update rule of BFE algorithm (e.g. as Figure 1 shows).

\subsection{Adaptive Extension of BFE Algorithm: AdaBFE}

\begin{algorithm}[H]
   \caption{Adaptive BFE of gradient change, the proposed algorithm in non-adaptive learning rate automation for stochastic optimization. See the text for details. We express Loss Function as $Loss=f(\theta)$. Default setting for error limit ratio for Binary Detection Learning Rate is $\epsilon = 0.001$, meaning one thousandth.}
   \label{alg:example}
\begin{algorithmic}
\STATE Initialize learning rate $\eta_i$ (e.g. 0.001)
   \STATE Initialize $\epsilon_{v,i}$ (e.g. 0.001)
   \STATE Initialize $\epsilon_{c,i} > \epsilon_{v,i}$
   \STATE Initialize parameter vector $\theta_0$
   \STATE Initialize time-step $t=0$
   \WHILE{$\theta_{t,i}$ not converged}
     \STATE $t=t+1$
     \IF{$\epsilon_{c,i} \geq \epsilon_{v,i}$}
       \WHILE{$\epsilon_{c,i} \geq \epsilon_{v,i}$}
         \STATE $g_i = \frac{\partial f(\theta)}{\partial \theta_{t,i}}$
         \STATE $\theta^{*}_{t,i} = \theta_{t,i} - \eta \cdot g_i$
         \STATE $g^{*}_i = \frac{\partial f(\theta)}{\partial \theta^{*}_{t,i}} $
         \STATE $\epsilon_{c,i} = arctan(abs((g^*_i-g_i)/(1+g^*_i \cdot g_i)))$
         \STATE $\epsilon_{v,i}= 1$ or any other values or functions
         \STATE $\eta_i = \frac{\eta_i}{2}$
       \ENDWHILE
       \STATE $\eta_i = 2\eta_i$
       \STATE $\theta_{t,i} = \theta^{*}_{t,i}$
     \ELSE
       \WHILE{$\epsilon_{c,i} < \epsilon_{v,i}$}
         \STATE $g_i = \frac{\partial f(\theta)}{\partial \theta_{t,i}}$
         \STATE $\theta^{*}_{t,i} = \theta_{t,i} - \eta \cdot g_i$
         \STATE $g^{*}_i = \frac{\partial f(\theta)}{\partial \theta^{*}_{t,i}} $
         \STATE $\epsilon_{c,i} = arctan(abs((g^*_i-g_i)/(1+g^*_i \cdot g_i)))$
         \STATE $\epsilon_{v,i}= 1$ or any other values or functions
         \STATE $\eta_i = 2\eta_i$
       \ENDWHILE
       \STATE $\eta_i = \frac{\eta_i}{2}$
       \STATE $\theta_{t,i} = \theta^{*}_{t,i}$
     \ENDIF
   \ENDWHILE
   \RETURN $\theta_{t}$ per dimension (Resulting Optimized Parameters)

\end{algorithmic}
\end{algorithm}

Based on the BFE of gradient change, we can extend it to the corresponding adaptive version. Instead of using a global learning rate, different learning rate $\eta_i$ in each dimension is updated through the criterion that comparing $\epsilon_{c,i}$ and $\epsilon_{v,i}$ in the corresponding dimension. Theoretically, the adaptive method enables to track and map loss landscape respectively for each dimension so that it can converge to a local minimum along a lowest energy-potential path if thinking in the perspective of physics.

\begin{figure}[ht]
\vskip 0.2in
\begin{center}
\centerline{\includegraphics[width=\columnwidth]{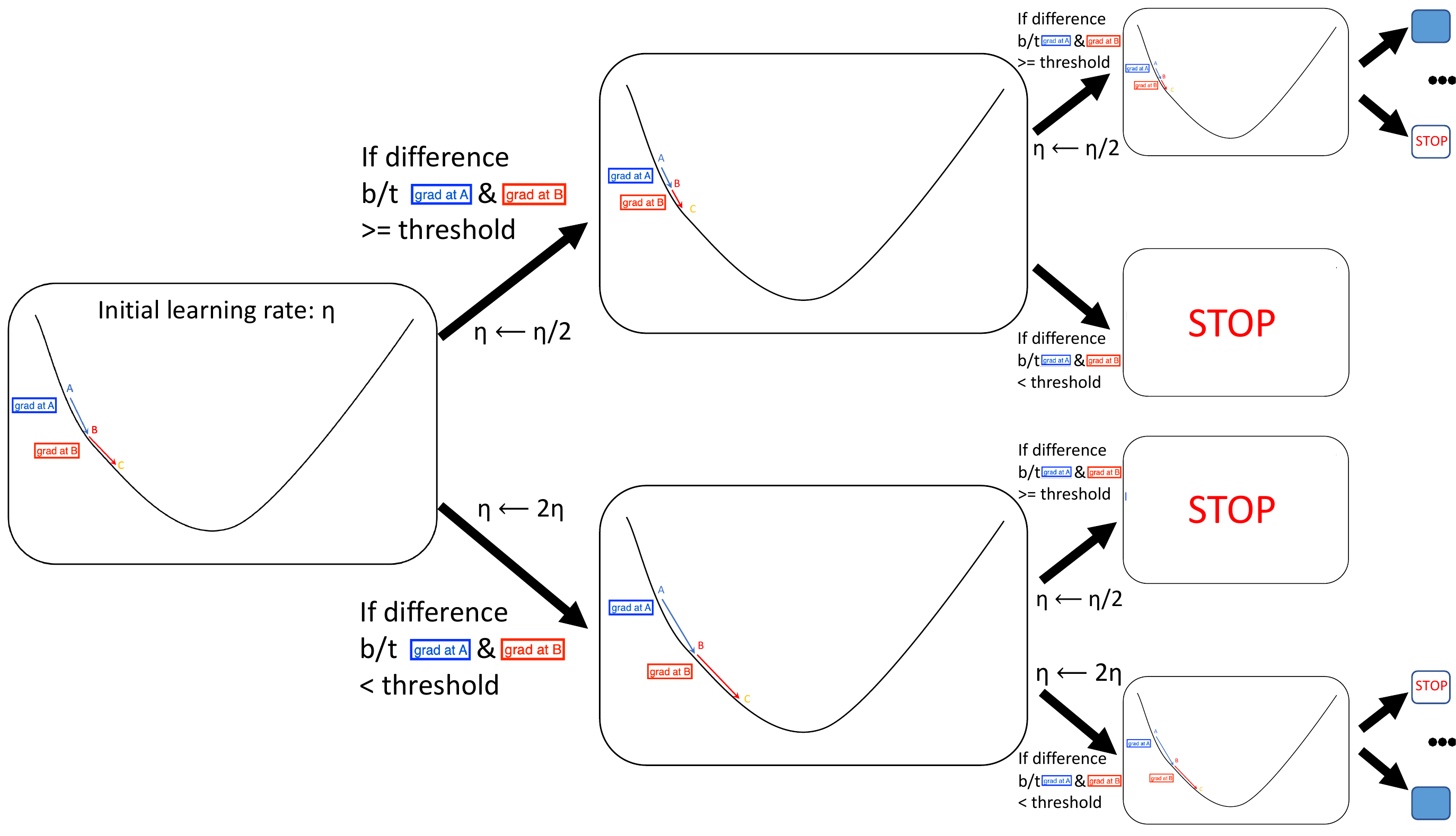}}
\caption{The update repeating process of BFE of gradient change algorithm, showing the iteration for a single time step's learning rate determination}
\label{icml-historical}
\end{center}
\vskip -0.2in
\end{figure}

Figure 9 demonstrated the comparison of loss variations between BFE, BFE of gradient change, the adaptive BFE of gradient change, the SGD and Adam algorithm for the univariate linear regression problem. Although Adam algorithm usually enable to make the optimization converge faster than SGD in the high-dimensional datasets, which is though different in the low-dimensional (e.g. 1D or 2D) regression \cite{gitman2018convergence}. For instance, the SGD algorithm could make the loss converge faster than Adam in the univariate linear regression, as Figure 9 shows. Meanwhile, it shows that the BFE of gradient change (default setting, e.g. $\epsilon_v$ = 1 degree) converges faster than the original BFE method, and the adaptive BFE of gradient change converges fastest compared with the other ones. The upper and lower panels in Figure 9 respectively reveal the loss decrease by using batch size of 128 and 512. The left, middle and right columns respectively reveal different time-step ranges. The original BFE started to speed up to converge during the $100^{th}-300^{th}$ time-step, and the adaptive BFE seems to become a little bit conservative during such a time-step range, which is probably because the adaptive BFE is trying to track the loss decrease carefully for a potential local minimum. Please note that, each time-step represents one iteration or one time-step of running one batch.

\begin{figure}[ht]
\vskip 0.2in
\begin{center}
\centerline{\includegraphics[width=\columnwidth]{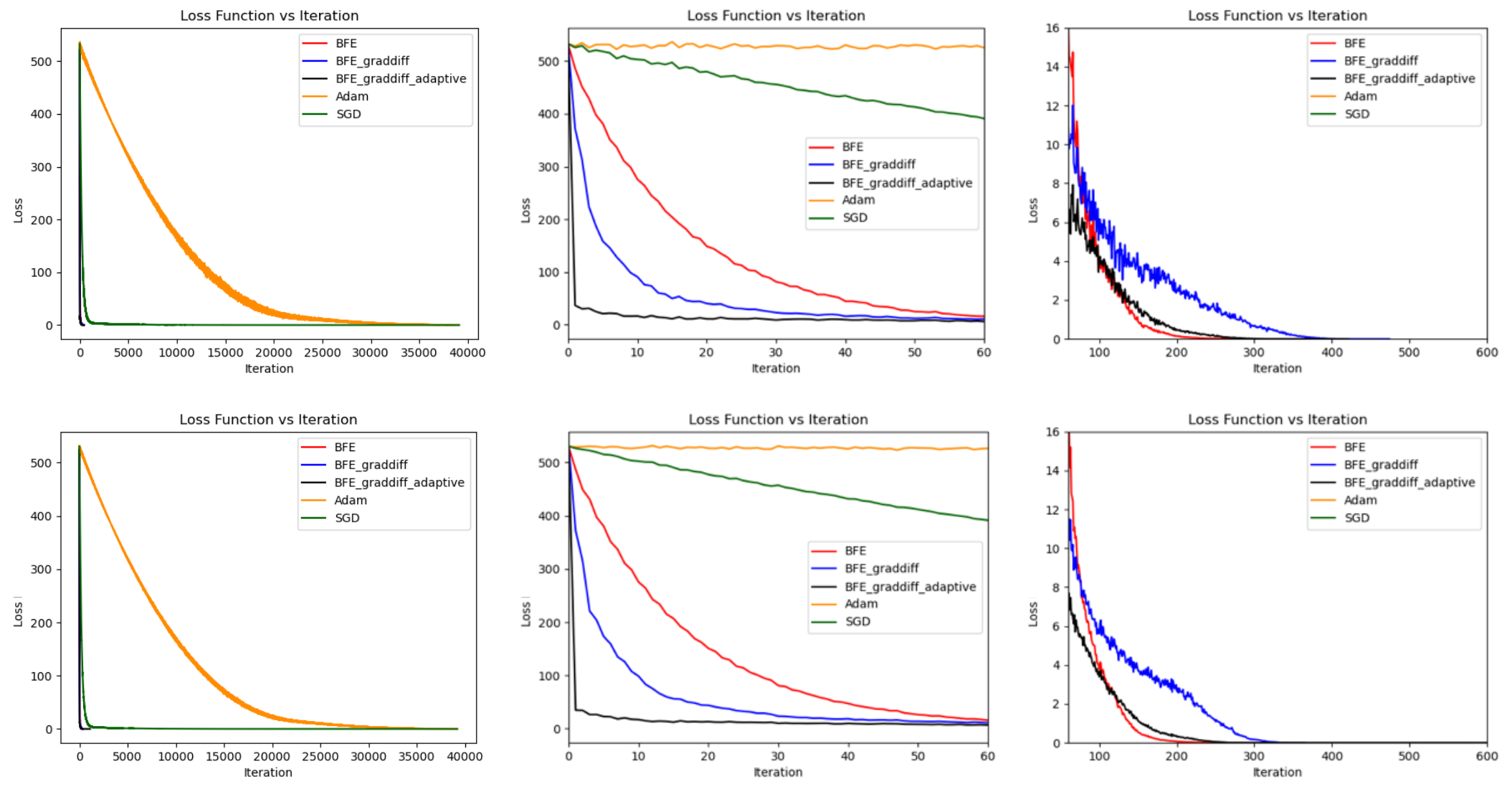}}
\caption{The loss variation with iterations over the whole dataset, which compares the BFE, BFE of gradient change with non-adaptive and adaptive versions. The left panels reveal the results during the first 60 time-steps, and the right panel shows the results during $60^{th}$ to $1000^{th}$ time-steps. The upper panels are the results with 128 batch size and the lower are 512 batch size.}
\label{icml-historical}
\end{center}
\vskip -0.2in
\end{figure}

In general, comparing different versions of BFE algorithm, the loss decrease with iterations were plotted in Figure 9. With the same setting of the common hyperparameters and initializations, the BFE of gradient change with non-adaptive and adaptive versions can outperform the original BFE algorithm during the beginning of loss values decrease, but the original BFE may enable the loss converges accelerate later. Investigating the patterns of different versions of BFE algorithm can help us better understand the multi-dimensional distribution in the parameter space.

\begin{algorithm}[tb]
   \caption{BFE (zoom-in part only), the proposed algorithm in non-adaptive learning rate automation for stochastic optimization. See the text for details. We express Loss Function as $Loss=f(\theta)$. Default setting for error limit ratio for Binary Detection Learning Rate is $\epsilon = 0.001$, indicating one thousandth.}
   \label{alg:example}
\begin{algorithmic}
   \STATE Initialize learning rate $\eta$ (e.g. 0.001)
   \STATE Initialize $\epsilon_v$ (e.g. 0.001)
   \STATE Initialize $\epsilon_c > \epsilon_v$
   \STATE Initialize parameter vector $\theta_0$
   \STATE Initialize time-step $t=0$
   \WHILE{$\theta_t$ not converged}
     \STATE $t=t+1$
     \STATE $\eta=\eta_0$ or $2\eta_0$
     \STATE assert or let $\epsilon_c \geq \epsilon_v$
       \WHILE{$\epsilon_c \geq \epsilon_v$}
         \STATE $\theta^{*}_{t} = \theta_t - \eta \frac{\partial f(\theta)}{\partial \theta_{t}}$
         \STATE $\theta^{+}_{t} = \theta_t - \frac{\eta}{2} \frac{\partial f(\theta)}{\partial \theta_{t}}$
         \STATE $\theta^{'}_{t} = \theta^{+}_{t} - \frac{\eta}{2} \frac{\partial f(\theta)}{\partial \theta^{+}_{t}}$
         \STATE $Loss1 = [f(\theta)]_{\theta^{*}_{t}}$ $ $ $ $ (loss value at $\theta^{*}_{t}$)
         \STATE $Loss2 = [f(\theta)]_{\theta^{'}_{t}}$ $ $ $ $ (loss value at $\theta^{'}_{t}$)
         \STATE $\epsilon_c = |Loss2-Loss1|$
         \STATE $\epsilon_v= 0.5 \cdot (|Loss2|+|Loss1|) \cdot \epsilon$ or $min(|Loss2|\cdot \epsilon, |Loss1|\cdot \epsilon)$ or any other predefined factor
         \STATE $\eta = \frac{\eta}{2}$
       \ENDWHILE
       \STATE $\theta_{t} = \theta^{+}_{t}$
   \ENDWHILE
   \RETURN $\theta_{t}$ (Resulting Optimized Parameters)

\end{algorithmic}
\end{algorithm}

In practice, it is also feasible not to update the learning rate in each time-step. Instead, we can update it every certain amount of time-steps if the loss landscape does not sharply fluctuate in the parameter space. Furthermore, under some circumstances, there is no need to consider the “while $\epsilon_{comp}<\epsilon_{val}$ do” loop or the zoom-out process. Instead, only the “while $\epsilon_{comp}\geq\epsilon_{val}$ do” loop or zoom-in process is considered, the algorithm of which is shown in Algorithm 4. This zoom-in only algorithm version could speed up the optimization for some specific types of loss landscapes. In addition, the binary shrink and amplification of the learning rate can also be extended to multiple-times shrink and amplification such as ternary, quinary or decimal, which decrease and increase the learning rate by three times, five times or ten times, for suiting the different optimization efficiency. In addition, more modified version of BFE related algorithms are listed in but not limited to the Appendix.

\section{Summary}

In conclusion, a new approach in learning rate automation for stochastic optimization called Binary Forward Exploration has been developed. This approach explores the potential loss landscape ahead of time for each step moving forward via calculating the variation of loss or gradient. The BFE and its variants (e.g. BFE of gradient change, adaptive BFE of gradient change) is able to determine the learning rate more suitable in tracking the loss landscape. Compared to other algorithms, the optimization performs much better especially for the cold start period of gradient descent process, which made it possible to avoid of using warm-up strategy in some circumstances. Most importantly, the newly proposed algorithms provided a different perspective to study the stochastic optimization process and investigate the relationship between the learning rate and the loss landscape in the parameter space. In addition, the new algorithms combined the relatively low time and computational complexity of the first-order optimization methods combined with the advantages of the more efficient second-order optimization methods, the latter of which actually provided us the second-order (variational) information of the loss landscape. The BFE algorithm is not only able to optimize the parameters for machine learning problems, but has also potential in solving the ODE/PDE equations or updating time-dependent equations in the physical scientific problems \cite{Cao2020a, Cao2020b, cao2019interaction, cao2020influence, cao2021using, Cao2017, cao2013multifluid, cao2014seasonal, cao20153d, cao20163d, cao2017diurnal, cao2018diurnal, cao2018magnetosphere, cao2021asymmetric}.

\section*{Acknowledgements}

The main idea of my this work was raised during 2016-2017, when I was pursuing my CSE degree at Georgia Tech. The methodology and philosophy originally derive from my 2013 modeling work, as mentioned in the paper. The writing of this manuscript has been completed recently.

\bibliography{main}
\bibliographystyle{icml2021}

\newpage

\begin{appendices}

\end{appendices}

\begin{algorithm}[H]
   \caption{Binary Forward Exploration (BFE), the proposed algorithm in non-adaptive learning rate automation for stochastic optimization. See the text for details. Loss function is denoted as $Loss=f(\theta)$. Default setting for error limit ratio for Binary Detection Learning Rate is $\epsilon = 0.001$, meaning one thousandth.}
   \label{alg:example}
\begin{algorithmic}
   \STATE Initialize learning rate $\eta$ (e.g. 0.001)
   \STATE Initialize $\epsilon_v$ (e.g. 0.001)
   \STATE Initialize $\epsilon_c > \epsilon_v$
   \STATE Initialize parameter vector $\theta_0$
   \STATE Initialize time-step $t=0$
   \WHILE{$\theta_t$ not converged}
     \STATE $t=t+1$
     \IF{$\epsilon_c \geq \epsilon_v$}
       \WHILE{$\epsilon_c \geq \epsilon_v$}
         \STATE $\theta^{*}_{t} = \theta_t - \eta \frac{\partial f(\theta)}{\partial \theta_{t}}$
         \STATE $\theta^{+}_{t} = \theta_t - \frac{\eta}{2} \frac{\partial f(\theta)}{\partial \theta_{t}}$
         \STATE $\theta^{'}_{t} = \theta^{+}_{t} - \frac{\eta}{2} \frac{\partial f(\theta)}{\partial \theta^{+}_{t}}$
         \STATE $Loss1 = [f(\theta)]_{\theta^{*}_{t}}$ $ $ $ $ (loss value at $\theta^{*}_{t}$)
         \STATE $Loss2 = [f(\theta)]_{\theta^{'}_{t}}$ $ $ $ $ (loss value at $\theta^{'}_{t}$)
         \STATE $\epsilon_c = |Loss2-Loss1|$
         \STATE $\epsilon_v= 0.5 \cdot (|Loss2|+|Loss1|) \cdot \epsilon$ or $min(|Loss2|\cdot \epsilon, |Loss1|\cdot \epsilon)$ or any other predefined factor or functions, e.g. decay with epochs
         \STATE $\eta = \frac{\eta}{2}$
       \ENDWHILE
       \STATE $\eta = 2\eta$
       \STATE $\theta_{t} = \theta^{*}_{t}$
     \ELSE
       \WHILE{$\epsilon_c < \epsilon_v$}
         \STATE $\theta^{+}_{t} = \theta_t - \eta \frac{\partial f(\theta)}{\partial \theta_{t}}$
         \STATE $\theta^{'}_{t} = \theta^{+}_t - \eta \frac{\partial f(\theta)}{\partial \theta^{+}_{t}}$
         \STATE $\theta^{*}_{t} = \theta_{t} - 2\eta \frac{\partial f(\theta)}{\partial \theta_{t}}$
         \STATE $Loss1 = [f(\theta)]_{\theta^{'}_{t}}$ $ $ $ $ (loss value at $\theta^{'}_{t}$)
         \STATE $Loss2 = [f(\theta)]_{\theta^{*}_{t}}$ $ $ $ $ (loss value at $\theta^{*}_{t}$)
         
         \STATE $\epsilon_c = |Loss2-Loss1|$
         \STATE $\epsilon_v= 0.5 \cdot (|Loss2|+|Loss1|) \cdot \epsilon$ or $min(|Loss2|\cdot \epsilon, |Loss1|\cdot \epsilon)$ or any other factor or functions
         \STATE $\eta = 2\eta$
       \ENDWHILE
       \STATE $\eta = \frac{\eta}{2}$
       \STATE $\theta_{t} = \theta^{+}_{t}$
     \ENDIF
   \ENDWHILE
   \RETURN $\theta_{t}$ (Resulting Optimized Parameters)
   
\end{algorithmic}
\end{algorithm}

\begin{algorithm}[H]
   \caption{BFE of gradient change, the proposed algorithm in non-adaptive learning rate automation for stochastic optimization. See the text for details. We express Loss Function as $Loss=f(\theta)$. Default setting for error limit ratio for Binary Detection Learning Rate is $\epsilon = 0.001$, meaning one thousandth.}
   \label{alg:example}
\begin{algorithmic}
   \STATE Initialize learning rate $\eta$ (e.g. 0.001)
   \STATE Initialize $\epsilon_v$ (e.g. 0.001)
   \STATE Initialize $\epsilon_c > \epsilon_v$
   \STATE Initialize parameter vector $\theta_0$
   \STATE Initialize time-step $t=0$
   \WHILE{$\theta_t$ not converged}
     \STATE $t=t+1$
     \IF{$\epsilon_c \geq \epsilon_v$}
       \WHILE{$\epsilon_c \geq \epsilon_v$}
         \STATE $g_i = \frac{\partial f(\theta)}{\partial \theta_{t,i}}$
         \STATE $\theta^{*}_{t,i} = \theta_{t,i} - \eta \cdot g_i$
         \STATE $g^{*}_i = \frac{\partial f(\theta)}{\partial \theta^{*}_{t,i}} $
         \STATE $\epsilon_i = arctan(abs((g^*_i-g_i)/(1+g^*_i \cdot g_i)))$
         \STATE $\epsilon_c = max(\epsilon_i)$
         \STATE $\epsilon_v= 1$ or any other values
         \STATE $\eta = \frac{\eta}{2}$
       \ENDWHILE
       \STATE $\eta = 2\eta$
       \STATE $\theta_{t,i} = \theta^{*}_{t,i}$
     \ELSE
       \WHILE{$\epsilon_c < \epsilon_v$}
         
         \STATE $g_i = \frac{\partial f(\theta)}{\partial \theta_{t,i}}$
         \STATE $\theta^{*}_{t,i} = \theta_{t,i} - \eta \cdot g_i$
         \STATE $g^{*}_i = \frac{\partial f(\theta)}{\partial \theta^{*}_{t,i}} $
         \STATE $\epsilon_i = arctan(abs((g^*_i-g_i)/(1+g^*_i \cdot g_i)))$
         \STATE $\epsilon_c = max(\epsilon_i)$
         \STATE $\epsilon_v= 1$ or any other values
         \STATE $\eta = 2\eta$
       \ENDWHILE
       \STATE $\eta = \frac{\eta}{4}$
       \STATE $\theta_{t,i} = \theta_{t,i} - \eta \cdot g_i$
     \ENDIF
   \ENDWHILE
   \RETURN $\theta_{t}$ (Resulting Optimized Parameters)
\end{algorithmic}
\end{algorithm}

\begin{algorithm}[H]
   \caption{Adaptive BFE of gradient change, the proposed algorithm in non-adaptive learning rate automation for stochastic optimization. See the text for details. We express Loss Function as $Loss=f(\theta)$. Default setting for error limit ratio for Binary Detection Learning Rate is $\epsilon = 0.001$, meaning one thousandth.}
   \label{alg:example}
\begin{algorithmic}
\STATE Initialize learning rate $\eta_i$ (e.g. 0.001)
   \STATE Initialize $\epsilon_{v,i}$ (e.g. 0.001)
   \STATE Initialize $\epsilon_{c,i} > \epsilon_{v,i}$
   \STATE Initialize parameter vector $\theta_0$
   \STATE Initialize time-step $t=0$
   \WHILE{$\theta_{t,i}$ not converged}
     \STATE $t=t+1$
     \IF{$\epsilon_{c,i} \geq \epsilon_{v,i}$}
       \WHILE{$\epsilon_{c,i} \geq \epsilon_{v,i}$}
         \STATE $\eta_i = \frac{\eta_i}{2}$
         \STATE $g_i = \frac{\partial f(\theta)}{\partial \theta_{t,i}}$
         \STATE $\theta^{*}_{t,i} = \theta_{t,i} - \eta \cdot g_i$
         \STATE $g^{*}_i = \frac{\partial f(\theta)}{\partial \theta^{*}_{t,i}} $
         \STATE $\epsilon_{c,i} = arctan(abs((g^*_i-g_i)/(1+g^*_i \cdot g_i)))$
         \STATE $\epsilon_{v,i}= 1$ or any other values or functions
       \ENDWHILE
       \STATE $\theta_{t,i} = \theta^{*}_{t,i}$
     \ELSE
       \WHILE{$\epsilon_{c,i} < \epsilon_{v,i}$}
         \STATE $g_i = \frac{\partial f(\theta)}{\partial \theta_{t,i}}$
         \STATE $\theta^{*}_{t,i} = \theta_{t,i} - \eta \cdot g_i$
         \STATE $g^{*}_i = \frac{\partial f(\theta)}{\partial \theta^{*}_{t,i}} $
         \STATE $\epsilon_{c,i} = arctan(abs((g^*_i-g_i)/(1+g^*_i \cdot g_i)))$
         \STATE $\epsilon_{v,i}= 1$ or any other values or functions
         \STATE $\eta_i = 2\eta_i$
       \ENDWHILE
       \STATE $\eta_i = \frac{\eta_i}{2}$
       \STATE $\theta_{t,i} = \theta^{*}_{t,i}$
     \ENDIF
   \ENDWHILE
   \RETURN $\theta_{t}$ per dimension (Resulting Optimized Parameters)

\end{algorithmic}
\end{algorithm}

\end{document}